%% file: template.tex
\documentclass[cameraready]{Interspeech}
\graphicspath{ {./figures/} }
\usepackage{hyperref}
\usepackage{booktabs,multirow,multicol,cite,makecell,enumitem,amsmath}
\usepackage[normalem]{ulem}
\useunder{\uline}{\ul}{}

\usepackage{caption}
\title{LLM-Based Multi-Reference Evaluation for\\Efficient and Robust Assessment of Phrase Break Annotations}

\author[affiliation={1,2}, equalcontribution, correspondingauthor]{Younghan}{Park}
\author[affiliation={1}, equalcontribution]{Hoyeon}{Lee}
\author[affiliation={1,3}]{Hawon}{Jeong}
\author[affiliation={1}]{Jong-Hwan}{Kim}

\address{
    $^1$NAVER Cloud, South Korea \\
    $^2$Yonsei University, South Korea \qquad
    $^3$KAIST AI, South Korea
}

\email{younghanpark@yonsei.ac.kr, yeon.lee@navercorp.com}


\begin{document}

\maketitle

\begin{abstract}
Reliable evaluation of phrase break annotations is crucial, as subtle variations in prosodic boundaries directly affect the clarity and naturalness of speech. However, existing approaches exhibit major limitations: single-reference evaluation assumes a unique gold phrasing for an utterance despite multiple valid phrasings, while human judgment, though flexible, is labor-intensive and unscalable. To address these, we propose LLM-based Multi-Reference Evaluation (LMRE) for phrase break annotations that models the one-to-many nature of prosodic phrasing and generates multiple valid phrasings from minimal demonstrations. On a Korean testbed of 1,356 annotations covering five strategies, LMRE shows stronger alignment with human judgment than single-reference evaluation in both acceptance behavior and score correlation. Our findings demonstrate that LMRE effectively achieves both scalability and multi-reference support, highlighting the potential of LLMs for evaluation in the speech domain.
\end{abstract}

\input{contents/1-introduction}
\input{contents/2-proposed-method}
\input{contents/3-experimental-setup}
\input{contents/4-experimental-results}

\section{Conclusion}
We introduce LMRE, a framework that models the one-to-many nature of prosodic phrasing using LLM-generated references, thereby addressing the limitations of conventional approaches for evaluating phrase break annotations.
By accounting for multiple valid annotations, LMRE reduces the under-acceptance problem inherent in single-reference evaluation and achieves stronger alignment with human judgment.
It consistently improves alignment with human scores while requiring only minimal human annotation effort.
This consistency across evaluation metrics, utterance lengths, and reference generation settings highlights its potential as a scalable and reliable solution for prosodic evaluation in TTS.

\section{Generative AI Use Disclosure}
Generative AI (ChatGPT by OpenAI and Claude by Anthropic) was used solely for grammatical correction and minor language refinement of the manuscript. The tool did not contribute to the scientific content, analysis, results, or conclusions of this work. All authors take full responsibility for the content of the paper.

\bibliographystyle{IEEEtran}
\bibliography{mybib}

\end{document}

%% file: contents/1-introduction.tex
\section{Introduction}

Prosodic information plays a central role in structuring and understanding speech \cite{frazier2006prosodic_phrasing}.
In particular, rhythmic cues, such as pauses or junctures, strongly influence how listeners segment continuous speech into meaningful units \cite{cutler1992rhythmic_cues}.
As the same principle holds for text-to-speech (TTS) systems \cite{pouw2025ip_in_tts}, explicit front-end modules for phrase break prediction have been widely studied \cite{kim2006pbp_for_korean, klimkov2017pbp_longform, futamata2021pbp, vadapalli2016investigation, mishra2015intonational, lee2025synthetic}.
Such modules have been shown to improve naturalness, intelligibility, and clarity of synthesized speech \cite{yang2023duration_aware_pi, hwang2023pausespeech, he2025prosodyfm}.

\begin{figure}[t]
\centering
\includegraphics[width=\columnwidth]{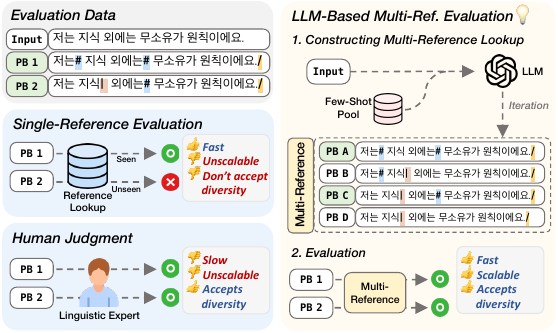}
\caption{Overview of approaches for evaluating phrase break prediction systems: conventional methods (left, blue box) and LLM-based multi-reference evaluation (LMRE; right).}
\label{fig:main}
\end{figure}

Since progress in phrase break prediction depends on reliable evaluation, the design of an efficient and robust evaluation framework is essential.
Two conventional approaches are commonly used to evaluate phrase break annotations (Figure~\ref{fig:main}, left).
In \textit{Single-Reference Evaluation}~\cite{futamata2021pbp, vadapalli2016investigation, liu2020modeling, mishra2015intonational, braunschweiler2016pause_prediction, lee2025synthetic, vadapalli2014continuous_pbp}, a phrase break annotation is compared against a unique gold annotation retrieved from a one-to-one lookup that is manually constructed by human annotators.
Once constructed, the reference lookup can be reused to automatically evaluate new systems.
However, when evaluation needs to cover previously unseen utterances, the lookup must be manually expanded, which is costly and unscalable.
A further limitation arises from the nature of prosodic phrasing, where multiple valid phrasings may exist for the same utterance \cite{brierley2007pbp_eval, hamlaoui2015flexible_ip, borise2022flexible_ip_verb_height}.
As single-reference evaluation assumes a one-to-one correspondence between an utterance and its phrasing, it poses a risk of incorrectly rejecting alternative yet valid phrasings that differ from the unique gold phrasing.

Another conventional approach is to conduct \textit{Human Judgment} \cite{klimkov2017pbp_longform, lee2025synthetic, parlikar2011grammar, brierley2007pbp_eval}.
Unlike single-reference evaluation, this method does not assume a single gold phrasing, thereby accounting for multiple valid phrasings.
However, it does not mitigate the scalability and efficiency issues of single-reference evaluation, since all phrasing variants must be independently assessed.
To address these limitations, \cite{wang2023assessing_pb_esl} explored the use of a large language model (LLM) as a judge for evaluating phrase break annotations.
However, although LLM-based judging is considered promising in many NLP tasks \cite{liu2023geval, zheng2023judging, kim2024prometheus, kocmi2023llm_translation}, its effectiveness remains limited for phrase break evaluation, as it struggles to capture subtle prosodic variations in speech.

To this end, we introduce \textit{LLM-based Multi-Reference Evaluation} (LMRE; Figure~\ref{fig:main}, right) for phrase break annotations, an evaluation framework that explicitly models the one-to-many nature of prosodic phrasing.
However, obtaining multiple valid annotations for each utterance through manual labeling is expensive and time-consuming.
Motivated by recent studies on using LLMs to generate diverse synthetic references for open-ended natural language generation \cite{tang2024not_all_metrics, liu2024summarize_llms_reference, zeng2024multiple_references, zhang2025reviseval}, we leverage LLMs to produce multiple plausible prosodic break annotations for TTS.
Unlike black-box LLM-as-a-judge approaches, LMRE constructs a reusable multi-reference lookup, enabling deterministic and reproducible evaluation.

We validate LMRE on a curated Korean testbed of 1,356 phrase break annotations spanning five different annotation strategies and eleven distinct configurations.
The results show that LMRE addresses the one-to-one assumption inherent in single-reference evaluation and substantially reduces the under-acceptance of valid annotations.
Moreover, it consistently achieves higher correlation with human scores across models and configurations, even with a small demonstration pool, demonstrating strong generalization to unseen utterances without additional human effort.
Overall, by mitigating the trade-off between scalability and multi-reference support, LMRE offers a scalable and reliable framework for prosodic evaluation in TTS systems.
We summarize our contributions as follows:
\begin{itemize}[leftmargin=1em]
\item We propose LMRE, a framework for evaluating phrase break annotations that overcomes the limitations of conventional methods by modeling the one-to-many nature of prosodic phrasing through LLM-generated reference annotations.
\item We construct a Korean testbed of 1,356 phrase break annotations spanning five annotation strategies and eleven configurations from prior literature for comprehensive validation under heterogeneous quality conditions.
\item Our findings show that LMRE enables more faithful phrase break evaluation with substantially improved alignment with human judgments over single-reference evaluation, while remaining fully automatic and scalable.
\end{itemize}

%% file: contents/2-proposed-method.tex
\section{Proposed Method}

\subsection{Evaluating Phrase Break Annotations}

Let an utterance $W = \{w_1, w_2, \ldots, w_n\}$ be a sequence of words, where $n$ is the number of words.
A phrase break annotation is defined as a corresponding sequence of phrase break labels $B = \{b_1, b_2, \ldots, b_n\}$.
Such annotations may be produced by humans or trained models, including LLMs.
In this work, we use a language-agnostic annotation scheme, drawing on prior studies on prosodic boundary labeling \cite{lee2025synthetic, kim2006pbp_for_korean, braunschweiler2016pause_prediction, silbervarod2022perception_breaks, klimkov2017pbp_longform}.
Specifically, we define $b_i\in\{\text{NB}, \text{AP}, \text{IP}, \text{SB}\}$, where $\text{NB} = \text{no boundary}$, $\text{AP} = \text{accent phrase boundary}$, $\text{IP} = \text{intonation phrase boundary}$, and $\text{SB} = \text{end-of-sentence boundary}$.

Let $B_h$ denote a hypothesis annotation to be evaluated.
An evaluator $\mathcal{F}$ determines whether to accept or reject a hypothesis annotation given an utterance, i.e., $\mathcal{F}(W, B_h)\in\{0, 1\}$, where 1 indicates acceptance and 0 indicates rejection.

\noindent \textbf{Single-Reference Evaluation}
A single-reference evaluator $\mathcal{F}_S$ assumes a one-to-one correspondence between an input utterance and its gold annotation, retrieving a reference annotation $B_r=R(W)$ for input $W$, where $R$ is a single-reference lookup.
It then computes a similarity metric $f$ between $B_h$ and $B_r$ and makes a decision using threshold $\theta$:
\[
\mathcal{F}_S(W, B_h|R) = \begin{cases}
1, & \text{if } f(B_h, B_r=R(W)) > \theta, \\
0, & \text{otherwise.}
\end{cases}
\]
In practice, $f$ is most commonly instantiated as either exact match (EM) or $F_1$ score.
When using $F_1$, an appropriate threshold $\theta$ must be chosen depending on the application domain.
Since $R$ is defined only for a fixed set of input texts, it must be manually expanded whenever new utterances are introduced, limiting efficiency and scalability.

\noindent \textbf{Human Judgment}  
Alternatively, evaluation can be performed directly by human judges.
Formally, a human judge $\mathcal{F}_H$ gives a binary decision on a hypothesis annotation $B_h$ for an input $W$, i.e., $\mathcal{F}_H(W, B_h) \in \{0, 1\}$.
Unlike single-reference evaluation, this approach does not assume a unique gold phrasing for an utterance, as humans can recognize distinct yet acceptable phrasing patterns.
However, this approach requires annotators with linguistic expertise and substantial manual effort, limiting its use in large-scale evaluation settings.

\subsection{LLM-Based Multi-Reference Evaluation (LMRE)}

\textbf{Multi-Reference Evaluation}
To explicitly model the one-to-many correspondence between utterance and valid phrasings, we define a multi-reference lookup $\mathbf{R}$ that maps an input $W$ to a set of reference annotations, i.e., $\mathbf{R}(W)=\{B^{(1)}_r, B^{(2)}_r, \ldots, B^{(k)}_r\}$.
The evaluator accepts a hypothesis if it achieves a similarity score greater than threshold $\theta$ with at least one reference in $\mathbf{R}(W)$:\footnote{Using average pooling is non-trivial, as the highest-scoring annotation under average pooling is not necessarily acceptable.}
\[
\mathcal{F}_M(W, B_h|\mathbf{R}) = \begin{cases}
1, & \text{if } \max_{B^{(i)}_r\in\mathbf{R}(W)}f(B_h, B^{(i)}_r) > \theta, \\
0, & \text{otherwise.}
\end{cases}
\]

\noindent \textbf{LLM-Based Reference Diversification}
We leverage LLMs to efficiently construct the multi-reference lookup $\mathbf{R} = \text{LLM}(P_\text{FS}, N_{\text{FS}}, N_{\text{iter}})$.
Here, $P_\text{FS}$ is the annotation pool for few-shot sampling, $N_{\text{FS}}$ is the number of demonstrations per query, and $N_{\text{iter}}$ is the number of LLM queries per utterance.
At each iteration, we resample a new set of few-shot examples from $P_\text{FS}$, while following the prompt format from \cite{lee2025synthetic}.
We fix the temperature at 0.0 for deterministic and reproducible outputs.

\noindent \textbf{Implementation Detail}
To mitigate spurious or noisy references, we retain only annotations generated more than $N_{\text{iter}}/10$ times across $N_{\text{iter}}$ iterations.
To reduce inference latency and cost while preserving reference quality, we employ batch prompting \cite{cheng2023batch} with a batch size of 32.\footnote{Both the threshold and batch size were selected based on preliminary experiments, balancing reference diversity, quality, and efficiency.}

%% file: contents/3-experimental-setup.tex
\input{tables/stats}

\section{Experimental Setup}

\begin{figure*}[t]
\centering
\includegraphics[width=\textwidth]{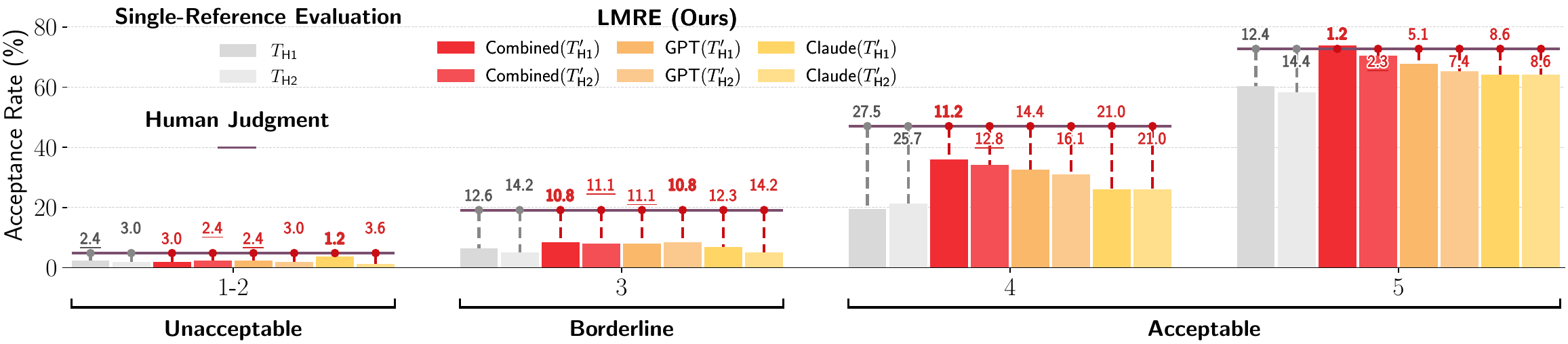}
\caption{
Acceptance rates (\%) of each evaluation method across hypothesis annotations grouped by their average human score ($f=\textnormal{EM}$, $|P_\textnormal{FS}|=128$, $N_\textnormal{iter}=20$).
Numbers above the horizontal line (Human Judgment) indicate the absolute difference between each evaluation method and human judgment.
\textbf{Bold} and \uline{underline} denote the smallest (best) and second-smallest differences.
}
\label{fig:exp1}
\end{figure*}

\subsection{Hypothesis Annotations}

We curate a Korean testbed comprising 1,356 phrase break annotations, spanning five annotation strategies and eleven configurations.
While LMRE is not language-specific, we focus on Korean to enable strict quality control under all configurations.
To ensure coverage of various syntactic and prosodic structures, we collect utterances across diverse domains, including conversations, GPS navigation, and news.
As annotation quality varies in practice, evaluating systems under heterogeneous quality conditions is essential to ensure robustness.
We therefore collect annotations using five different strategies, including rule-based baselines, human-written annotations, and LLM-generated annotations:
\begin{itemize}[leftmargin=1em]
    \item \textit{AP-Only}: rule-based annotations with no NB or IP labels, i.e., $B=\{\text{AP}, \cdots, \text{AP}, \text{SB}\}$, regardless of input utterance
    \item \textit{Comma-IP}: annotations identical to AP-Only, except that an IP appears whenever a word ends with a comma
    \item \textit{Audio-driven}: annotations derived from human speech recordings of the texts \cite{futamata2021pbp, lee2023transfer_pbp, vadapalli2014continuous_pbp}
    \item \textit{Text-driven}: annotations created by linguistic experts relying solely on the texts \cite{lee2025synthetic}
    \item \textit{Synthetic}: synthetic annotations generated by LLMs \cite{lee2025synthetic}
\end{itemize}

For audio-driven and text-driven annotations, we obtain annotations from recordings of three native Korean speakers and two linguistic experts, respectively, following the protocol from \cite{futamata2021pbp, lee2023transfer_pbp, lee2025synthetic}.
For synthetic annotations, we prompt LLMs (\texttt{\small gpt-4.1-mini} and \texttt{\small gpt-4.1}) using text-driven annotations from each of the two human experts as $P_\text{FS}$, resulting in four configurations.
As shown in Table~\ref{tab:stats}, the testbed consists of multiple phrasings per utterance, spanning diverse quality levels.
We further categorize utterances by length as follows: short ($<7$ words), medium (7--10 words), and long ($\geq11$ words).

\subsection{Evaluation Methods}

\noindent \textbf{Single-Reference Evaluation}
For single-reference evaluation, we use two text-driven reference lookups, $R \in \{ T_\text{H1}, T_\text{H2} \}$, provided by two human experts, H1 and H2.
To ensure reliability, the references were cross-validated through independent review by the annotators and the authors.

\noindent \textbf{Human Judgment}
We collect human judgment data from four independent judges who were not involved in testbed curation, thereby avoiding potential bias.
The judges had access to an internal annotation-to-speech tool, which was used as a supplementary aid for reliable assessment.
Two judges provided binary judgments, while the other two used a 1–5 scale to enable fine-grained analysis:
\begin{itemize}[leftmargin=1em]
\item 5: The annotation is good in terms of naturalness or clarity.
\item 4: The annotation hardly affects naturalness or clarity.
\item 3: The annotation somewhat affects naturalness or clarity.
\item 2: The annotation harms naturalness or clarity.
\item 1: The annotation strongly disrupts naturalness or clarity.
\end{itemize}

\noindent \textbf{LLM-Based Multi-Reference Evaluation (LMRE)}
For the few-shot pool $P_\text{FS}$, we use text-driven annotations from the same two experts, H1 and H2.
These sets are disjoint from $T_\text{H1}$ and $T_\text{H2}$ and are denoted as $T'_\text{H1}$ and $T'_\text{H2}$ to prevent trivial copying and to evaluate generalization to unseen utterances.
We set $N_\text{FS}$ to half of $|P_\text{FS}|$, which maximizes the number of distinct few-shot combinations that can be formed.
For constructing the multi-reference lookup, we experiment with three LLM settings: using only \texttt{\small gpt-4.1} (``GPT''), only \texttt{\small claude-sonnet-4-20250514} (``Claude''), and combining the reference sets generated by both models into a unified lookup (``Combined'').
In the Combined setting, $N_\text{iter}$ is evenly split between the two models.
For fair comparison between single-model and Combined settings, we keep the total $N_\text{iter}$ identical across configurations unless otherwise specified.

\subsection{Metrics}

\textbf{Robustness to Reference Diversity}
We use acceptance rate to quantify the extent to which each evaluation method recognizes alternative valid annotations.
The acceptance rate is defined as the proportion of hypothesis annotations that are accepted by the evaluator, i.e., $\mathcal{F}(W, B_h) = 1$.

\noindent \textbf{Alignment with Human Judgment}
To evaluate alignment with human judgment, we compute the Pearson and Spearman correlation coefficients between the scores produced by each evaluation method and the corresponding five-scale human scores.
For the metric $f$, we employ EM and $F_1$ score, as they are standard metrics in evaluating phrase break annotations \cite{brierley2007pbp_eval, futamata2021pbp, lee2025synthetic, lee2023transfer_pbp, klimkov2017pbp_longform, vadapalli2014continuous_pbp}.

%% file: tables/stats.tex
\begin{table}[t]
\setlength{\tabcolsep}{4pt}
\centering
\scriptsize
\renewcommand{\arraystretch}{1.05}

\newcommand{\cpsep}{@{\hspace{1.5pt}}}

\begin{tabular}{@{}l c c@{\hspace{2pt}}c c@{\hspace{2pt}}c c@{\hspace{2pt}}c c@{\hspace{2pt}}c@{}}
\toprule
\multicolumn{2}{l}{\multirow{2}{*}{}} & \multicolumn{8}{c}{Subsets} \\
\cmidrule(lr){3-10}
\multicolumn{2}{l}{} & \multicolumn{2}{c}{Short} & \multicolumn{2}{c}{Medium} & \multicolumn{2}{c}{Long} & \multicolumn{2}{c}{All} \\
\midrule

\multicolumn{2}{c}{\# Utterances}
& \multicolumn{2}{c}{112}
& \multicolumn{2}{c}{120}
& \multicolumn{2}{c}{68}
& \multicolumn{2}{c}{300} \\
\midrule

\multirow{6}{*}[-0.2em]{\rotatebox[origin=c]{90}{\shortstack{\# Unique\\ Annotations}}}
 & 1 & 0   & (0.00\%)  & 5   & (0.82\%)  & 11  & (2.67\%)  & 16  & (1.18\%)  \\
 & 2 & 9   & (2.72\%)  & 75  & (12.23\%) & 65  & (15.78\%) & 149 & (10.99\%) \\
 & 3 & 56  & (16.92\%) & 168 & (27.41\%) & 97  & (23.54\%) & 321 & (23.67\%) \\
 & 4 & 152 & (45.92\%) & 265 & (43.23\%) & 196 & (47.57\%) & 613 & (45.21\%) \\
 & 5 & 114 & (34.44\%) & 100 & (16.31\%) & 43  & (10.44\%) & 257 & (18.95\%) \\

\cmidrule(l){2-10}

 & $\Sigma$
 & 331   & (100.0\%)
 & 613   & (100.0\%)
 & 412   & (100.0\%)
 & 1,356 & (100.0\%) \\

\bottomrule
\end{tabular}

\caption{Statistics of collected hypothesis annotations with distribution across 1--5 human score groups.}
\label{tab:stats}
\end{table}

%% file: contents/4-experimental-results.tex
\section{Experimental Results}

\input{tables/ccof}

\subsection{Comparison of Evaluation Methods}
\label{ssec:compare}

We compare acceptance behavior across three evaluation methods: (1) single-reference evaluation, (2) human judgment, and (3) LMRE.
To verify that the method appropriately rejects low-quality annotations and accepts high-quality ones, we group annotations into three categories based on the integer part of their average human scores: Unacceptable (1–2), Borderline (3), and Acceptable (4–5).
The results are summarized in Figure~\ref{fig:exp1}.

\subsubsection{Human Judgment}
As anticipated, higher human scores correspond to higher acceptance rates.
Annotations in the Unacceptable or Borderline groups are rarely accepted, whereas annotations in the Acceptable group are frequently accepted.
Annotations with a score of 3 exhibit an acceptance rate of 19.19\%, reflecting borderline cases in which evaluators may differ in their decisions.

\subsubsection{Single-Reference Evaluation}
In contrast to human judgment, acceptance rates under single-reference evaluation remain consistently lower across all score groups.
In the Unacceptable group, the average gap relative to human judgment is marginal (2.73\%), indicating its comparable performance in rejecting poor annotations.
However, this discrepancy becomes pronounced where multiple valid phrasings exist, particularly in the Acceptable group.
Specifically, the average gap reaches 26.59\% in score group 4 and 13.42\% in score group 5.
These results suggest that single-reference evaluation systematically under-accepts valid annotations that differ from the unique reference, highlighting its structural limitation.

\subsubsection{LLM-Based Multi-Reference Evaluation (LMRE)}
LMRE narrows the gap to human judgment across all score groups, reducing the under-acceptance observed in single-reference evaluation.
The improvement is most pronounced in the Acceptable group, where acceptance rates under LMRE closely track human judgment, reducing the average gap to 7.04\% in the Combined setting.
Notably, the gap narrows even further to 1.75\% for score group 5.
These findings demonstrate that LMRE effectively mitigates the structural limitation of single-reference evaluation.

Importantly, this improvement is consistent across different few-shot pools.
The differences between using $T'_\text{H1}$ and $T'_\text{H2}$ as the few-shot pool are negligible across all score groups, regardless of the LLM used to construct the reference lookup.
Furthermore, these results are obtained with a relatively small few-shot pool ($|P_\text{FS}|=128$), suggesting that LMRE generalizes well to unseen utterances even from limited demonstrations.

\subsection{Ablation Study}
\label{ssec:ccof}

\begin{figure}[t]
\centering
\includegraphics[width=\columnwidth]{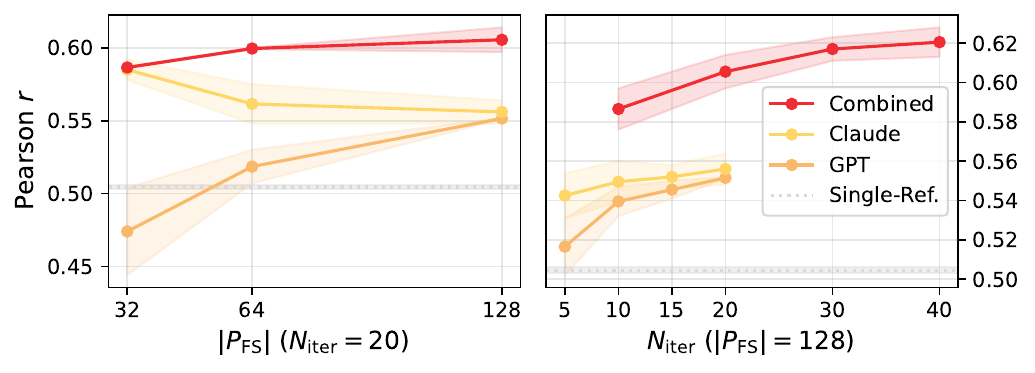}
\caption{Pearson correlation as a function of few-shot pool size $|P_\textnormal{FS}|$ and the number of iterations $N_\textnormal{iter}$. Shading shows the absolute difference between using $T'_\textnormal{H1}$ and $T'_\textnormal{H2}$ as $P_\textnormal{FS}$.}
\label{fig:exp2}
\end{figure}

In this section, we examine how LMRE's hyperparameters affect its alignment with human perception.
The results are summarized in Table~\ref{tab:ccof} and Figure~\ref{fig:exp2}.

\noindent \textbf{Improved Alignment with Human Scores}
Across most conditions, LMRE achieves higher correlations with human scores than single-reference evaluation.
In particular, in the Combined$^\dagger$ setting, the correlations between human scores and multi-reference $F_1$ reach $r=0.621$ and $\rho=0.626$, which are notably higher than those obtained with the single-reference evaluation setting ($r=0.505$ and $\rho=0.497$).

\noindent \textbf{Evaluation Metric Choice}
Overall, $F_1$ shows stronger alignment with human judgment than EM.
As EM requires an exact sequence-level match with the reference, it is sensitive to minor prosodic variations that humans may consider acceptable.
In contrast, $F_1$ rewards partial overlap between annotations, accommodating subtle phrasing variations.
These characteristics are reflected in the empirical results.
In the single-reference setting on the All subset, $r$ increases from 0.416 with EM to 0.505 with $F_1$, and $\rho$ increases from 0.454 to 0.497.
The gap becomes larger under LMRE (Combined$^\dagger$), where $r$ increases from 0.530 with EM to 0.621 with $F_1$, and $\rho$ increases from 0.560 to 0.626.
These results indicate that both single-reference evaluation and LMRE benefit more when the metric rewards partial agreement rather than requiring exact matches.

\noindent \textbf{Trends across Utterance Length}
The relative gain of LMRE over single-reference evaluation increases with utterance length under EM.
Using Pearson correlation, the improvement from single-reference evaluation to LMRE (Combined$^\dagger$) grows from 0.132 (Short) and 0.090 (Medium) to 0.169 (Long), with a similar pattern observed for Spearman.
This trend is consistent with the intuition that longer utterances permit greater phrasing variability, which is better captured by multi-reference evaluation.
However, the gain does not increase and instead slightly decreases with length under $F_1$.
A possible explanation is that $F_1$ already gives partial credit for overlapping subsequences, leaving less room for additional improvement from considering alternative phrasings.
Nevertheless, LMRE consistently improves over single-reference evaluation across all length groups under both EM and $F_1$, despite the different gain patterns.

\noindent \textbf{Effect of Few-Shot Pool Size and Iterations}
Increasing the few-shot pool size leads to more noticeable improvements for GPT and the Combined setting, while Claude remains comparatively stable.
All variants of our method benefit from increasing the number of iterations, with performance improving consistently as $N_{\text{iter}}$ grows and gradually saturating at larger values.
In addition, the gap between using $T'_\text{H1}$ and $T'_\text{H2}$ narrows as either $|P_{\text{FS}}|$ or $N_{\text{iter}}$ increases, indicating greater robustness to inter-annotator variability at larger scales.

%% file: tables/ccof.tex
\setlength{\tabcolsep}{2pt}

\newcommand{\subfontsize}[1]{{\tiny #1}}

\newcolumntype{R}{>{\raggedleft\arraybackslash}p{0.77cm}}

\begin{table*}[t]
\centering
\scriptsize
\begin{tabular}{l c *{16}{R}}
\toprule
 & \multicolumn{1}{c}{} & \multicolumn{8}{c}{\textbf{Single-Reference / Multi-Reference EM}} & \multicolumn{8}{c}{\textbf{Single-Reference / Multi-Reference} \boldsymbol{$F_1$}} \\
 \cmidrule(lr){3-10}
 \cmidrule(lr){11-18}\\
 \addlinespace[-9pt]

 & \multicolumn{1}{c}{}
 & \multicolumn{2}{c}{Short} & \multicolumn{2}{c}{Medium} & \multicolumn{2}{c}{Long} & \multicolumn{2}{c}{\cellcolor[HTML]{EFEFEF}All}
 & \multicolumn{2}{c}{Short} & \multicolumn{2}{c}{Medium} & \multicolumn{2}{c}{Long} & \multicolumn{2}{c}{\cellcolor[HTML]{EFEFEF}All} \\
\multirow{-3}{*}{Ref.}
& \multicolumn{1}{c}{\multirow{-3}{*}{\boldsymbol{$|P_\text{FS}|$}}}
& \multicolumn{1}{c}{$r$} & \multicolumn{1}{c}{$\rho$}
& \multicolumn{1}{c}{$r$} & \multicolumn{1}{c}{$\rho$}
& \multicolumn{1}{c}{$r$} & \multicolumn{1}{c}{$\rho$}
& \multicolumn{1}{c}{\cellcolor[HTML]{EFEFEF}$r$} & \multicolumn{1}{c}{\cellcolor[HTML]{EFEFEF}$\rho$}
& \multicolumn{1}{c}{$r$} & \multicolumn{1}{c}{$\rho$}
& \multicolumn{1}{c}{$r$} & \multicolumn{1}{c}{$\rho$}
& \multicolumn{1}{c}{$r$} & \multicolumn{1}{c}{$\rho$}
& \multicolumn{1}{c}{\cellcolor[HTML]{EFEFEF}$r$} & \multicolumn{1}{c}{\cellcolor[HTML]{EFEFEF}$\rho$} \\ \midrule

\addlinespace[-1.5pt]
\multicolumn{18}{l}{\textit{Single-Reference Evaluation}} \\
\addlinespace[-1pt]
\cmidrule{1-18}

$T_{\text{H}\cdot}$
& -
& .55\subfontsize{ (.02)} & .61\subfontsize{ (.02)} & .39\subfontsize{ (.01)} & .42\subfontsize{ (.02)} & .29\subfontsize{ (.03)} & .31\subfontsize{ (.03)}
& \cellcolor[HTML]{EFEFEF}.42\subfontsize{ (.01)} & \cellcolor[HTML]{EFEFEF}.45\subfontsize{ (.01)}
& .53\subfontsize{ (.02)} & .53\subfontsize{ (.02)} & .53\subfontsize{ (.01)} & .53\subfontsize{ (.00)} & .56\subfontsize{ (.00)} & .53\subfontsize{ (.01)}
& \cellcolor[HTML]{EFEFEF}.50\subfontsize{ (.00)} & \cellcolor[HTML]{EFEFEF}.50\subfontsize{ (.01)} \\
\midrule

\addlinespace[-1.5pt]
\multicolumn{18}{l}{\textit{LMRE (Ours)}: $N_\text{iter} = 20$ by default; $^\dagger$ indicates $N_\text{iter} = 40$.} \\
\addlinespace[-1pt]
\cmidrule{1-18}

\multirow{3}{*}{$\text{GPT}(T'_{\text{H}\cdot})$}
& 32
& .62\subfontsize{ (.04)} & .65\subfontsize{ (.05)} & .38\subfontsize{ (.06)} & .39\subfontsize{ (.06)} & .28\subfontsize{ (.11)} & .29\subfontsize{ (.09)}
& \cellcolor[HTML]{EFEFEF}.42\subfontsize{ (.07)} & \cellcolor[HTML]{EFEFEF}.45\subfontsize{ (.06)}
& .62\subfontsize{ (.05)} & .64\subfontsize{ (.04)} & .48\subfontsize{ (.05)} & .51\subfontsize{ (.06)} & .41\subfontsize{ (.13)} & .44\subfontsize{ (.11)}
& \cellcolor[HTML]{EFEFEF}.47\subfontsize{ (.06)} & \cellcolor[HTML]{EFEFEF}.51\subfontsize{ (.05)} \\
& 64
& .64\subfontsize{ (.01)} & {\ul .67}\subfontsize{ (.01)} & .43\subfontsize{ (.03)} & .44\subfontsize{ (.03)} & .33\subfontsize{ (.02)} & .34\subfontsize{ (.03)}
& \cellcolor[HTML]{EFEFEF}.46\subfontsize{ (.01)} & \cellcolor[HTML]{EFEFEF}.49\subfontsize{ (.01)}
& .64\subfontsize{ (.00)} & .65\subfontsize{ (.01)} & .55\subfontsize{ (.06)} & .56\subfontsize{ (.06)} & .48\subfontsize{ (.02)} & .50\subfontsize{ (.02)}
& \cellcolor[HTML]{EFEFEF}.52\subfontsize{ (.02)} & \cellcolor[HTML]{EFEFEF}.54\subfontsize{ (.02)} \\
& 128
& {\ul .66}\subfontsize{ (.02)} & \textbf{.70}\subfontsize{ (.02)} & .44\subfontsize{ (.00)} & .46\subfontsize{ (.00)} & .36\subfontsize{ (.01)} & .38\subfontsize{ (.01)}
& \cellcolor[HTML]{EFEFEF}.48\subfontsize{ (.01)} & \cellcolor[HTML]{EFEFEF}.51\subfontsize{ (.01)}
& .65\subfontsize{ (.01)} & {\ul .68}\subfontsize{ (.02)} & .60\subfontsize{ (.02)} & .60\subfontsize{ (.01)} & .50\subfontsize{ (.04)} & .50\subfontsize{ (.04)}
& \cellcolor[HTML]{EFEFEF}.55\subfontsize{ (.00)} & \cellcolor[HTML]{EFEFEF}.57\subfontsize{ (.00)} \\
\cmidrule{1-18}

\multirow{3}{*}{$\text{Claude}(T'_{\text{H}\cdot})$}
& 32
& .60\subfontsize{ (.01)} & .62\subfontsize{ (.01)} & {\ul .48}\subfontsize{ (.01)} & .49\subfontsize{ (.01)} & \textbf{.47}\subfontsize{ (.01)} & \textbf{.50}\subfontsize{ (.01)}
& \cellcolor[HTML]{EFEFEF}{\ul .52}\subfontsize{ (.01)} & \cellcolor[HTML]{EFEFEF}{\ul .55}\subfontsize{ (.01)}
& .57\subfontsize{ (.01)} & .59\subfontsize{ (.00)} & .61\subfontsize{ (.03)} & .60\subfontsize{ (.03)} & \textbf{.69}\subfontsize{ (.04)} & \textbf{.66}\subfontsize{ (.02)}
& \cellcolor[HTML]{EFEFEF}.59\subfontsize{ (.01)} & \cellcolor[HTML]{EFEFEF}.59\subfontsize{ (.01)} \\
& 64
& .54\subfontsize{ (.09)} & .56\subfontsize{ (.07)} & .45\subfontsize{ (.03)} & .46\subfontsize{ (.03)} & .43\subfontsize{ (.02)} & .47\subfontsize{ (.02)}
& \cellcolor[HTML]{EFEFEF}.49\subfontsize{ (.03)} & \cellcolor[HTML]{EFEFEF}.51\subfontsize{ (.03)}
& .50\subfontsize{ (.10)} & .52\subfontsize{ (.08)} & .59\subfontsize{ (.01)} & .59\subfontsize{ (.00)} & .65\subfontsize{ (.00)} & {\ul .64}\subfontsize{ (.00)}
& \cellcolor[HTML]{EFEFEF}.56\subfontsize{ (.03)} & \cellcolor[HTML]{EFEFEF}.57\subfontsize{ (.03)} \\
& 128
& .60\subfontsize{ (.05)} & .63\subfontsize{ (.06)} & .42\subfontsize{ (.03)} & .44\subfontsize{ (.03)} & .37\subfontsize{ (.03)} & .40\subfontsize{ (.03)}
& \cellcolor[HTML]{EFEFEF}.46\subfontsize{ (.03)} & \cellcolor[HTML]{EFEFEF}.50\subfontsize{ (.04)}
& .55\subfontsize{ (.04)} & .57\subfontsize{ (.05)} & .57\subfontsize{ (.02)} & .56\subfontsize{ (.02)} & .63\subfontsize{ (.01)} & .60\subfontsize{ (.02)}
& \cellcolor[HTML]{EFEFEF}.56\subfontsize{ (.02)} & \cellcolor[HTML]{EFEFEF}.55\subfontsize{ (.03)} \\
\cmidrule{1-18}

\multirow{3}{*}{Combined}
& 32
& .64\subfontsize{ (.02)} & .65\subfontsize{ (.03)} & .47\subfontsize{ (.02)} & .49\subfontsize{ (.01)} & .45\subfontsize{ (.03)} & .47\subfontsize{ (.03)}
& \cellcolor[HTML]{EFEFEF}.51\subfontsize{ (.01)} & \cellcolor[HTML]{EFEFEF}.54\subfontsize{ (.01)}
& .63\subfontsize{ (.04)} & .64\subfontsize{ (.03)} & .61\subfontsize{ (.02)} & .61\subfontsize{ (.01)} & .65\subfontsize{ (.01)} & .62\subfontsize{ (.01)}
& \cellcolor[HTML]{EFEFEF}.59\subfontsize{ (.00)} & \cellcolor[HTML]{EFEFEF}.60\subfontsize{ (.00)} \\
& 64
& .64\subfontsize{ (.06)} & .65\subfontsize{ (.06)} & \textbf{.49}\subfontsize{ (.01)} & \textbf{.51}\subfontsize{ (.00)} & \textbf{.47}\subfontsize{ (.03)} & {\ul .49}\subfontsize{ (.03)}
& \cellcolor[HTML]{EFEFEF}{\ul .52}\subfontsize{ (.01)} & \cellcolor[HTML]{EFEFEF}{\ul .55}\subfontsize{ (.01)}
& .62\subfontsize{ (.07)} & .63\subfontsize{ (.08)} & {\ul .64}\subfontsize{ (.05)} & {\ul .63}\subfontsize{ (.05)} & {\ul .66}\subfontsize{ (.01)} & .63\subfontsize{ (.01)}
& \cellcolor[HTML]{EFEFEF}.60\subfontsize{ (.00)} & \cellcolor[HTML]{EFEFEF}.60\subfontsize{ (.01)} \\
& 128
& \textbf{.68}\subfontsize{ (.04)} & \textbf{.70}\subfontsize{ (.04)} & .47\subfontsize{ (.00)} & .49\subfontsize{ (.00)} & .45\subfontsize{ (.01)} & .47\subfontsize{ (.01)}
& \cellcolor[HTML]{EFEFEF}{\ul .52}\subfontsize{ (.01)} & \cellcolor[HTML]{EFEFEF}{\ul .55}\subfontsize{ (.01)}
& {\ul .66}\subfontsize{ (.05)} & {\ul .68}\subfontsize{ (.06)} & {\ul .64}\subfontsize{ (.01)} & {\ul .63}\subfontsize{ (.01)} & .65\subfontsize{ (.01)} & .63\subfontsize{ (.01)}
& \cellcolor[HTML]{EFEFEF}{\ul .61}\subfontsize{ (.02)} & \cellcolor[HTML]{EFEFEF}{\ul .61}\subfontsize{ (.01)} \\
\cmidrule{1-18}

\textbf{Combined\boldsymbol{$^\dagger$}}
& 128
& \textbf{.68}\subfontsize{ (.02)} & \textbf{.70}\subfontsize{ (.02)} & {\ul .48}\subfontsize{ (.01)} & {\ul .50}\subfontsize{ (.01)} & {\ul .46}\subfontsize{ (.01)} & .48\subfontsize{ (.01)}
& \cellcolor[HTML]{EFEFEF}\textbf{.53}\subfontsize{ (.01)} & \cellcolor[HTML]{EFEFEF}\textbf{.56}\subfontsize{ (.01)}
& \textbf{.67}\subfontsize{ (.02)} & \textbf{.69}\subfontsize{ (.02)} & \textbf{.65}\subfontsize{ (.01)} & \textbf{.64}\subfontsize{ (.01)} & {\ul .66}\subfontsize{ (.01)} & .63\subfontsize{ (.01)}
& \cellcolor[HTML]{EFEFEF}\textbf{.62}\subfontsize{ (.02)} & \cellcolor[HTML]{EFEFEF}\textbf{.63}\subfontsize{ (.01)} \\
\bottomrule
\end{tabular}

\caption{Pearson ($r$) and Spearman ($\rho$) correlation coefficients between automatic metrics (EM, $F_1$) and human scores across evaluation settings, with varying model choices and few-shot pool sizes.
Correlations are averaged over two annotators; values in parentheses show their absolute difference before averaging.
\textbf{Bold} indicates the best result, and \uline{underline} indicates the second-best result.
}
\label{tab:ccof}
\end{table*}